\documentclass[conference]{IEEEtran}
\IEEEoverridecommandlockouts
\usepackage{cite}
\usepackage{amsmath,amssymb,amsfonts}
\usepackage{algorithmic}
\usepackage{graphicx}
\usepackage{textcomp}
\usepackage{xcolor}
\usepackage{multirow}
\def\BibTeX{{\rm B\kern-.05em{\sc i\kern-.025em b}\kern-.08em
    T\kern-.1667em\lower.7ex\hbox{E}\kern-.125emX}}
\begin{document}

\title{A comparative evaluation of machine learning methods for robot navigation through human crowds\\
}
\author{\IEEEauthorblockN{Anastasia Gaydashenko}
\IEEEauthorblockA{\textit{Cisco, CTAO\thanks{Anastasia Gaydashenko conducted the research while at internship at JetBrains Research, St. Petersburg, Russia.}} \\
\textit{a.v.gaydashenko@gmail.com}}
\and
\IEEEauthorblockN{Daniel Kudenko}
\IEEEauthorblockA{\textit{Department of Computer Science}\\
\textit{University of York, UK} \\
\textit{JetBrains Research}\\
\textit{St. Petersburg, Russia} \\
\textit{daniel.kudenko@york.ac.uk}}
\and
\IEEEauthorblockN{Aleksei Shpilman}
\IEEEauthorblockA{\textit{JetBrains Research}\\
\textit{St. Petersburg, Russia} \\
\textit{alexey@shpilman.com}}
}

\maketitle

\begin{abstract}
 Robot navigation through crowds poses a difficult challenge to AI systems, since the methods should result in fast and efficient movement but at the same time are not allowed to compromise safety. Most approaches to date were focused on the combination of pathfinding algorithms with machine learning for pedestrian walking prediction. More recently, reinforcement learning techniques have been proposed in the research literature. In this paper, we perform a comparative evaluation of pathfinding/prediction and reinforcement learning approaches on a crowd movement dataset collected from surveillance videos taken at Grand Central Station in New York. The results demonstrate the strong superiority of state-of-the-art reinforcement learning approaches over pathfinding with state-of-the-art behaviour prediction techniques. 
\end{abstract}


\section{Introduction}
Robots are predicted to increasingly participate in everyday life of humans over the next decade, but so far most research applications deal with interactions between a robot and a single human. However, if robots are to perform general tasks such as pick-up and delivery in public spaces, they will need to be able to cope and interact with large crowds. This poses a specific challenge for navigation control, where robots are supposed to move fast and efficient, while not endangering the safety of humans walking about in the same location. 

Most work so far on robot navigation in environments with moving obstacles has been focused on a combination of pathfinding algorithms and human behaviour prediction \cite{bai2015intention}. However, more recently, reinforcement learning has been proposed for robot navigation through crowds \cite{chen2017decentralized}, but there has been no conclusive empirical comparison so far to pathfinding with prediction. In this paper, we present an evaluation of various pedestrian prediction techniques combined with the state-of-the-art pathfinding algorithm D* Lite \cite{d-star-lite}, and a recent reinforcement learning approach SA-CADRL \cite{chen2017socially}. 



We evaluate the different approaches on a dataset extracted from real crowd movements recorded by CCTV at Grand Central Station in New York \cite{yi2015understanding}. The evaluation criteria are speed of movement from start to end point, and number of collisions of the virtual robot with humans. Our empirical results demonstrate that the reinforcement learning approach results in one of the fastest robots, while significantly reducing the number of collisions with humans, as measured in simulation based on real human movement data. 

\section{Related work}

Research in the area of socially aware robot motion follows three main directions: behavior prediction, reinforcement learning, and behavior cloning. 

Most of the work so far has focused on predicting humans' trajectories and plan the robot movement accordingly \cite{ferrer2014proactive}. For the prediction of humans' trajectories, clustering and typical behavior extraction was shown to be effective (e.g. \cite{ferrer2014behavior}). More recently, Deep Learning techniques have been successfully applied to the prediction of individual and crowd motion \cite{yi2016pedestrian}, which is currently the best performing technique. For robot route planning with dynamic obstacles the approaches vary. The simplest approach is to use a pre-defined set of rules to move the robot \cite{sisbot2007human}. Another approach is to compute attracting and repelling steering forces based on current and (where available) predicted future positions of obstacles \cite{reynolds1999steering}. This steering technique has been extended with Markov Decision Processes to deal with uncertainty in the predictions \cite{bai2015intention}. Svenstrup et al. \cite{svenstrup2010trajectory} iteratively improve the current best path using a Rapidly-exploring Random Tree with obstacles represented as potential fields.

In a recent reinforcement learning approach \cite{chen2017decentralized}, the robot was trained in simulation to move optimally through a crowd. 

The third main approach to socially aware robot motion, behavior cloning, relies on the fact that humans can navigate in human crowds quite successfully and tries to imitate human behavior. This movement behaviour imitation problem has recently been tackled with inverse reinforcement learning, computing a reward function for crowd navigation from human crowd movement demonstrations, that was subsequently used to train a reinforcement learning agent \cite{kretzschmar2016socially}. 

In this paper, we are focusing on the first two approaches, and perform a comparative evaluation in a simulation based on real-world pedestrian crowd recordings. 

\section{Crowd movement dataset}

\begin{figure}[htpb]
    \centering
    \includegraphics[width=0.45\textwidth]{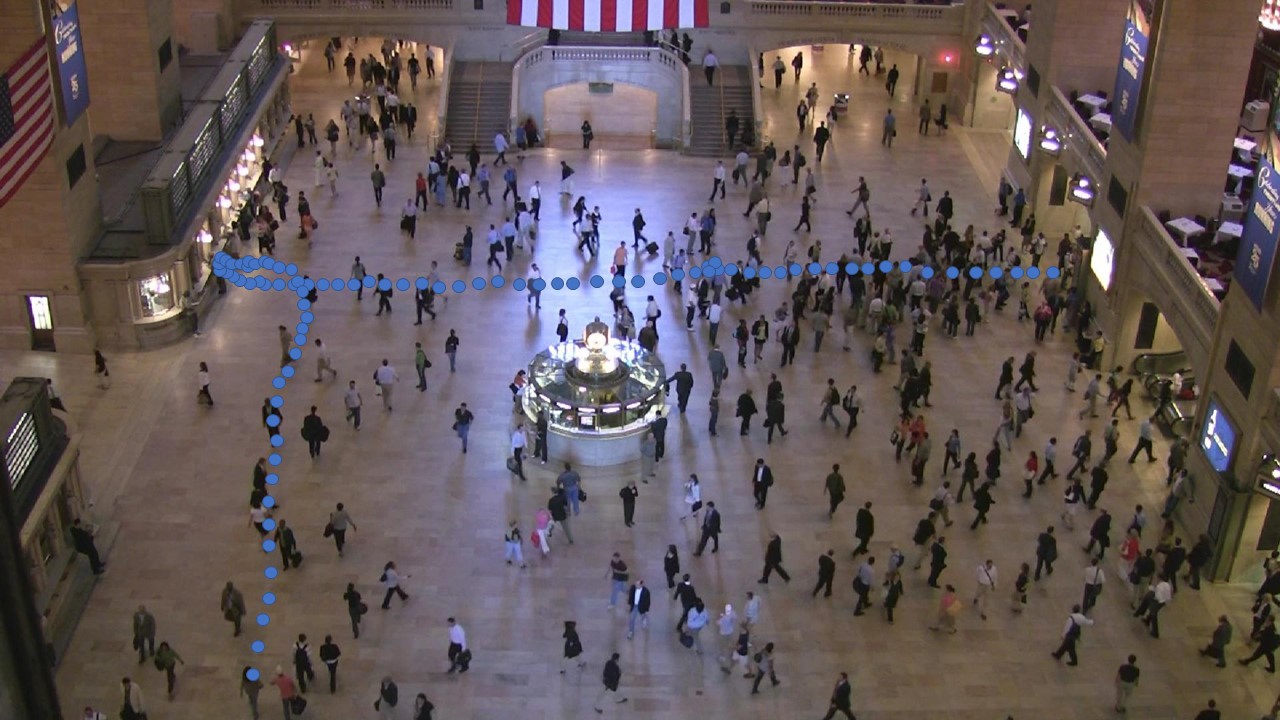}
    \caption{Example frame with sample trajectory for one person.}
    \label{fig:big_path}
\end{figure}

In order to simulate realistic crowd movements, we used a dataset obtained from surveillance video from Grand Central Station (New York, USA). The dataset consists of 6001 frames of camera images taken from the top, and 12684 labeled pedestrian trajectories as obtained by Yi et al. \cite{yi2015understanding}. An example frame with a single pedestrian trajectory is shown in Figure~\ref{fig:big_path}. The frames were sampled from a video of 4000 secs (i.e., a frequency is $1.5$ frames per second), and each frame has a resolution of $1920 \times 1080$. Trajectories were filtered by minimum length, where only those with at least 10 frames were retained. This left 12244 valid trajectories in the dataset with an average length of $37$ frames.

\begin{figure}[htpb]
    \centering
    \includegraphics[width=0.45\textwidth]{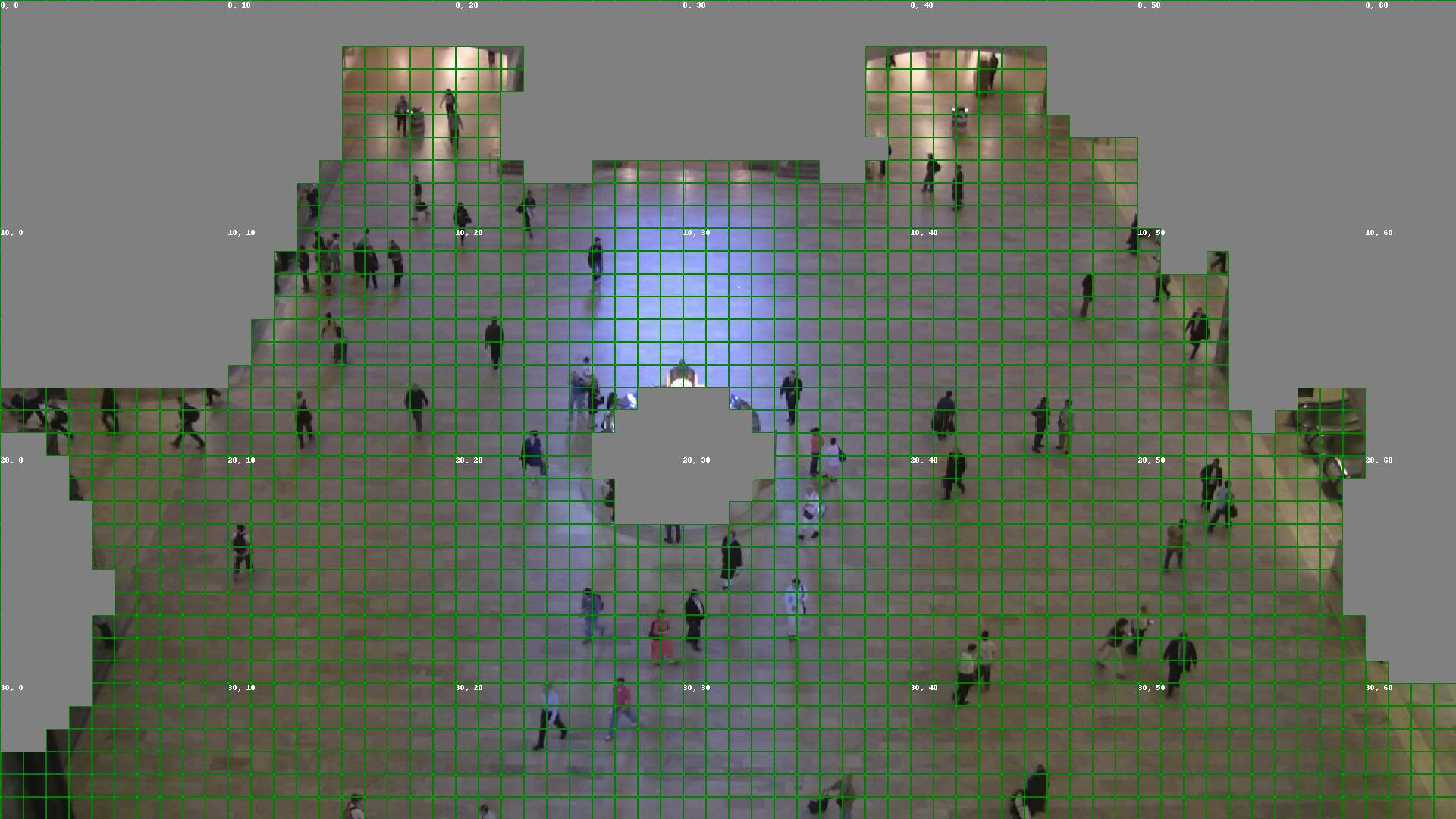}
    \caption{Overlaying a $64 \times 36$ grid.}
    \label{fig:grid}
\end{figure}

In a data pre-processing step, we overlaid the frames with a $64 \times 36$ grid, as shown in Figure~\ref{fig:grid}. This allows discretization of the pedestrian positions and reduces the size of the state space for the path planning. The grid resolution was chosen based on average human speed from frame to frame which are $35$ pixels for the horizontal coordinate and $30$ pixels for the vertical one.

\section{Path planning methods} \label{sec:prediction}

We employed two general approaches to the path planning problem. In the first approach pedestrians are considered as moving obstacles and the goal is to predict their positions over the next five time steps. These positions are then used as dynamic obstacles for the path planning algorithms such as D* Lite \cite{d-star-lite}, which will be described later. The second approach is to learn a robot movement policy via reinforcement learning in simulation based on the recorded pedestrian movements. We start with a discussion of prediction methods for pedestrian positions. 

\subsection{Pedestrian position prediction approach}

We compared several methods for position prediction in our study to demonstrate their impact on the robot policy. 

\subsubsection{Baseline prediction}

A very simple method to predict the future position of a pedestrian is to assume that the pedestrian can move into any neighboring grid square in a time step. All these squares are considered to be occupied by the pedestrian in the next time step when performing the path planning. 

\subsubsection{Random forest regression prediction}

A more sophisticated approach is to employ machine learning techniques for the prediction task. We chose Random Forest Regressor \cite{liaw2002classification} for this as a standard technique which has demonstrated robust performance in the literature. In this method, the prediction outputs are the five expected coordinates of the pedestrian over the next five time steps, relative to their original position. 

The state for which a prediction is made was represented using features based on the five previous grid positions of the pedestrian as follows:

\begin{itemize}
    \item Absolute coordinates (10 features).
    \item Coordinates relative to the previous point ($x$ and $y$ speed, 8 features).
    \item Speed (distance between neighboring points, 4 features) and average speed (1 feature).
    \item Acceleration (change in speed from one step to the next one, 3 features) and average acceleration (1 feature).
    \item Angle of movement for each step (4 features).
\end{itemize}

\begin{figure}
    \centering
    \includegraphics[width=0.49\textwidth]{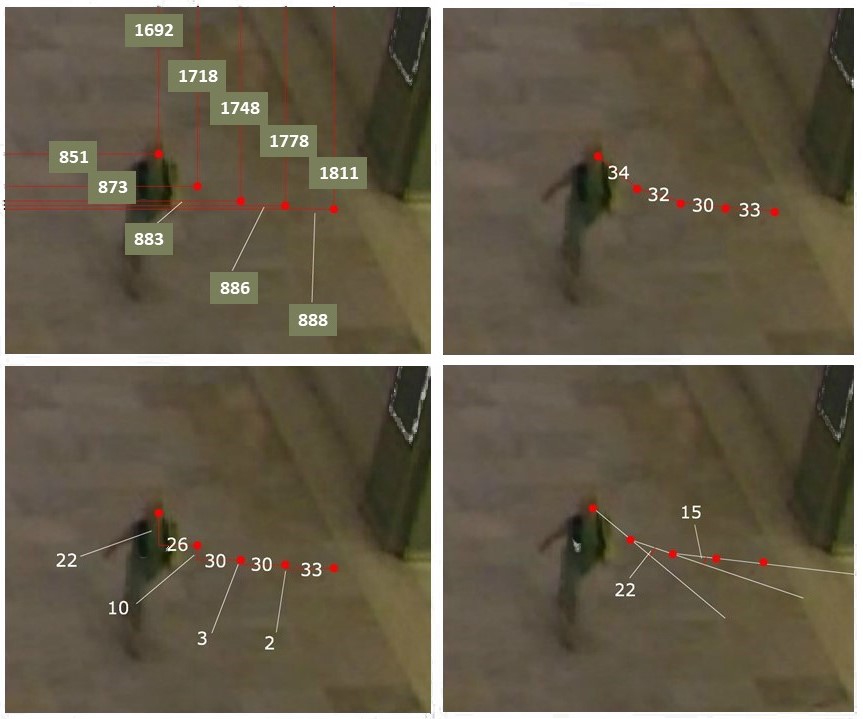}
    \caption{Features for Random Forest Regressor. The top left shows the x and y coordinates of the last five trajectory points. The top right picture shows the distance between neighbouring trajectory points. The bottom left picture shows the coordinates relative to the previous trajectory point. The bottom right picture shows the changes in movement angle between two time steps.}
    \label{fig:features}
\end{figure}

The features are illustrated on Fig. \ref{fig:features}

\subsubsection{Behavior-CNN prediction}

Behavior-CNN \cite{yi2016pedestrian} is a Neural Network architecture which was designed especially for movement prediction. It takes as input the embedded positions of the pedestrian over the last five time steps, and outputs the predicted positions over the next five time steps. 

These embedded positions are called a displacement volume and correspond to a three-dimensional matrix of size $X \times Y \times 10$, where $X$ and $Y$ denote the area size (in our case $X = 1920$ and $Y = 1080$). The third dimension corresponds to the distances between the pedestrian positions over the last 5 time steps, along the x and y dimension. This results in 10 values. The distance values are first normalized to be between 0 and 1 according to $X$ and $Y$ respectively, and then 1 is added to is to avoid too small numbers. More formally, the 10-dimensional vector is computed as given in equation~\ref{eq:bcnn}, where ($x_t$,$y_t$) is the $i$-th position of the pedestrian with $t=1$ being the time five steps ago, and $t=5$ the current time step:

\begin{equation} \label{eq:bcnn}
    [1 + \frac{x_5 - x_1}{X}, 1 + \frac{y_5 - y_1}{Y}, ..., 1 + \frac{x_5 - x_5}{X}, 1 + \frac{y_5 - y_5}{Y}]
\end{equation}

All other cells in the matrix are set to zero. The output the CNN has the same shape as the input and represents the predicted embedded positions over the next five time steps. 

For other technical details we refer the reader to the original paper.

\begin{figure*}
    \centering
    \includegraphics[width=1\textwidth]{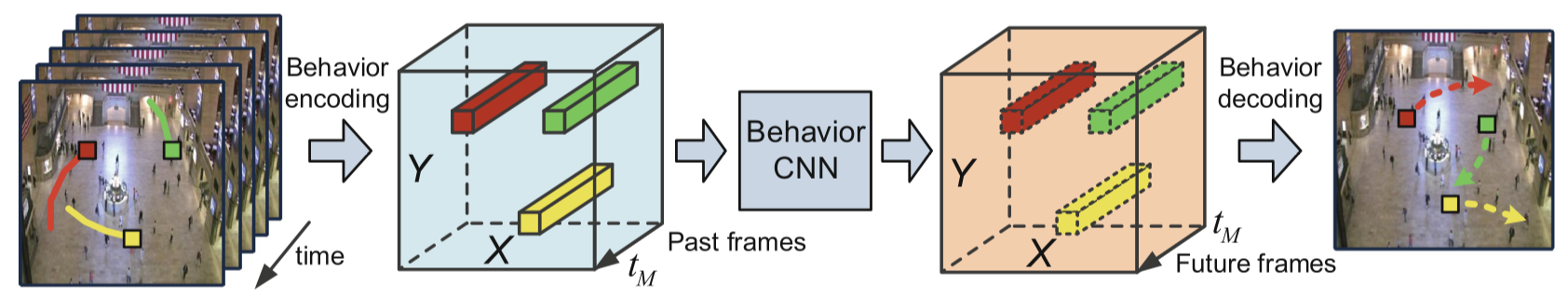}
    \caption{Behaviour-CNN framework \cite{yi2016pedestrian}.}
    \label{fig:beh-cnn}
\end{figure*}

\subsubsection{Path planning with the D* Lite algorithm} \label{sec:d-star-lite}

D* Lite \cite{d-star-lite} is a heuristic path planning algorithm that takes as input a start and goal position, and recalculates the current path at each step taking into account the predicted coordinates of obstacles over a number of time steps. The method uses the freespace assumption, i.e. it assumes that unknown space is empty unless the predictions tell otherwise. We refer the reader to the original paper for more details.

Realization in Python 3 was based on the java implementation GitHub \footnote{https://github.com/daniel-beard/DStarLiteJava}.



\subsection{Reinforcement learning approach}

\subsubsection{SA-CADRL}
A state-of-the-art RL approach to socially aware robot movement is SA-CADRL (Socially Aware Collision Avoidance with Deep Reinforcement Learning) \cite{chen2017socially} which aims to learn an optimal robot policy based on the observed state of the robot and nearby agents (i.e. pedestrians). During training, the robot receives a positive reward for reaching the destination and a negative reward for delays and collisions. The policy is represented as a neural network whose input is formed from state information on the robot and nearby agents (pedestrians) and that outputs the robot velocity. This state information includes the current position, size, and velocity of the agent/robot. 

Formally speaking, by using SA-CADRL architecture authors are trying to restore robot's policy $\pi$ such as expected value to goal would be minimal:

$$\arg \min_{\pi(s)} \mathbf{E}[t_g | \pi]$$

With the following restrictions:

\begin{itemize}
    \item $||p_t - \hat{p}_t||_2 \geq r + \hat{r}$ -- for any frame distance between robot and other agents should be greater than their radii sum.
    \item $p_{t_g} = p_g$ -- at the last frame robot should be in the goal coordinates.
    \item $p_t = p_{t-1} + \Delta t \cdot \pi(s)$ -- current robot's position is defined by previous position and chosen action (moving direction).
\end{itemize}

To evaluate this approach on the same trajectories and with the same metrics as the other approaches, we used the pre-trained network provided by the Chen et al. which can be found in the SA-CADRL project GitHub page \footnote{https://github.com/mfe7/cadrl\_ros}.

\begin{figure}
    \centering
    \includegraphics[width=0.45\textwidth]{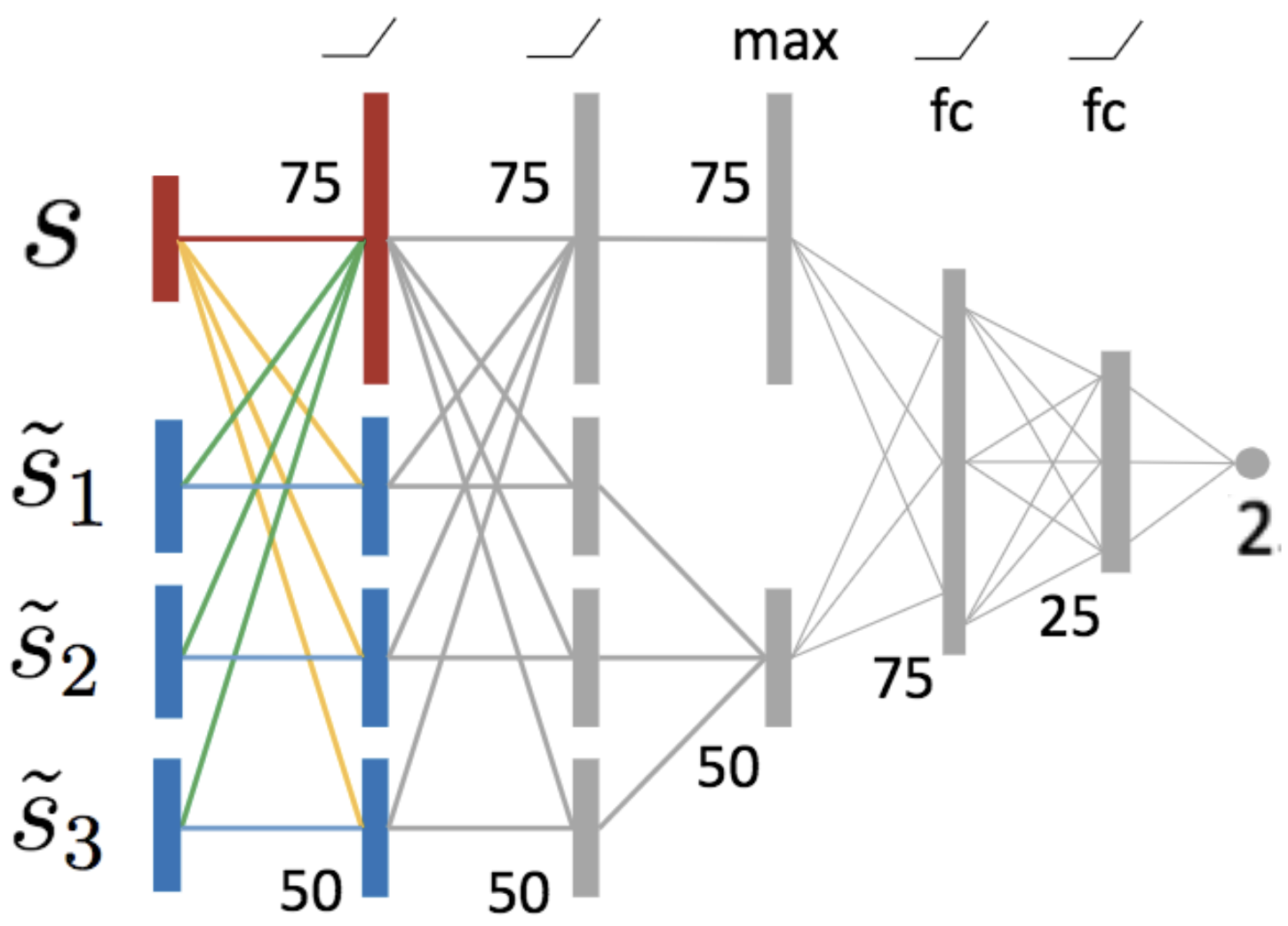}
    \caption{Original SA-CADRL network architecture \cite{chen2017socially}.}
    \label{fig:sa-cadrl}
\end{figure}

\subsubsection{Grid-SA-CADRL}

A limitation of the SA-CADRL approach is that the robot can't move backwards. We extended the SA-CADRL approach to provide the robot with these abilities. Furthermore, we modified SA-CADRL to work on a grid-environment rather than a coordinate space as the original approach. This means that rather than having a continuous movement action, the actions are made discrete corresponding to a move into one of the 8 neighboring grid squares or standing still. 

\begin{figure}[htpb]
    \centering
    \includegraphics[width=0.45\textwidth]{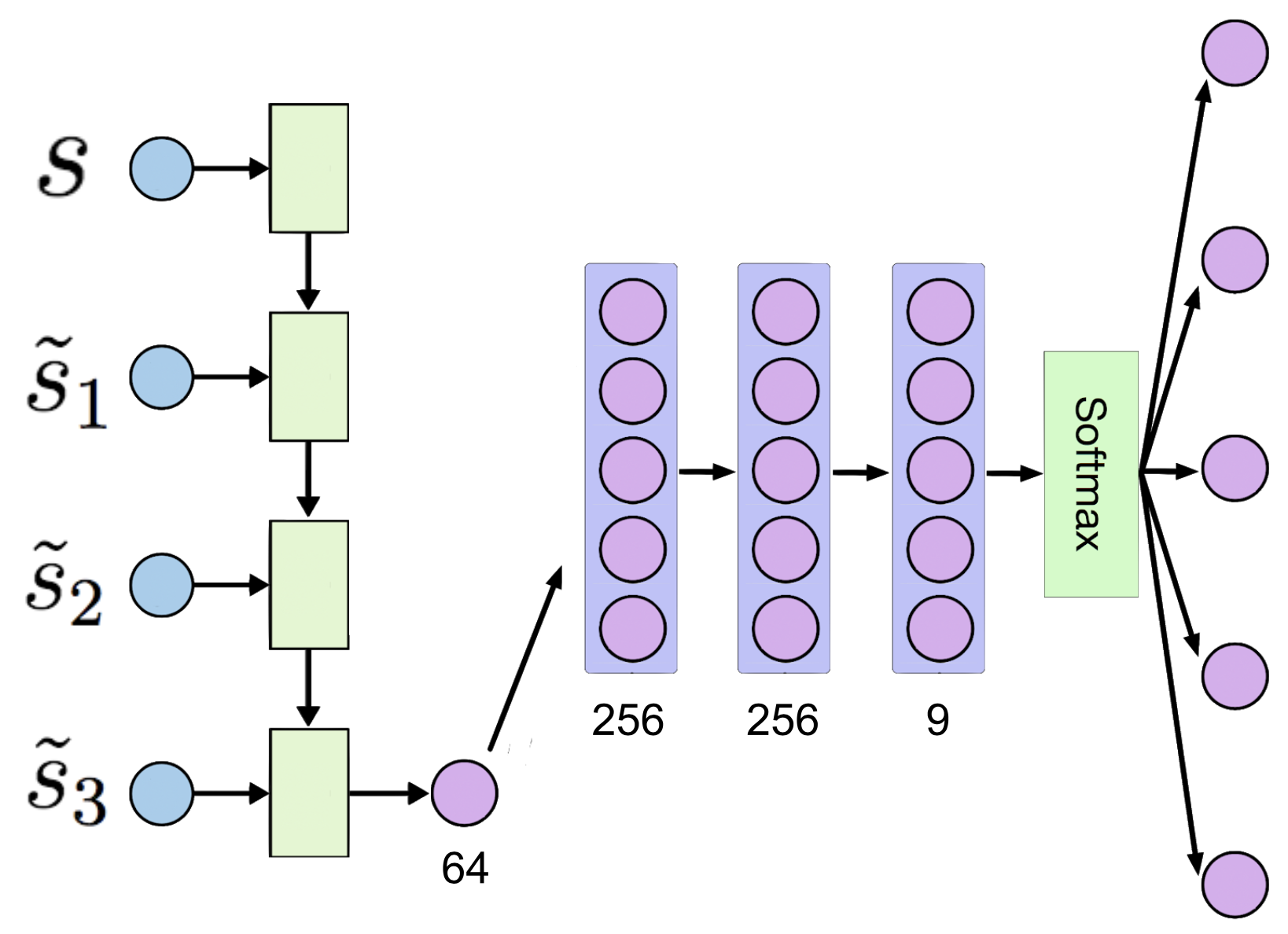}
    \caption{Grid-SA-CADRL network architecture.}
    \label{fig:network}
\end{figure}

The resulting network architecture is shown in Figure~\ref{fig:network}. It takes the robots and three nearby pedestrians' states as inputs into an LSTM layer (size 64), followed by 3 fully connected layers with the last one outputting probabilities of actions through a softmax transformation. The robot then choose the action with the highest probability.

We trained the Grid-SA-CADRL network for 1000 episodes, taking the pre-trained SA-CADRL network as a starting point. In each training episode we randomly picked a start and goal position for the robot, and a random starting time from the recordings. The pedestrians in the simulation walk according to the recording data, and an episode ends as soon as the robot has reached the goal position, or the recording has reached its end. 

\section{Evaluation}

We first present the evaluation of the prediction method accuracy, followed by the evaluation of the path planning approaches. 

\subsection{Evaluation of prediction methods}

We chose the Normalized Mean Square Error (NMSE, see equation below) to evaluate the pedestrian position prediction methods presented in Section~\ref{sec:prediction}, and optimize the hyper-parameters. 

\begin{equation}
NMSE = \frac{1}{n}\sum^{n}_{i=1}{\sqrt{(\frac{x_i - \hat{x_i}}{X})^2 + (\frac{y_i - \hat{y_i}}{Y})^2}}
\end{equation}

Here $x_i$ and $y_i$ are ground truth values, $x_i$ and $y_i$ are predicted. $X$ and $Y$ are upper limits for $x_i$ and $y_i$ respectively, and in our case set to $X = 1920$ and $Y = 1080$.

We randomly split the data set of pedestrian trajectories into 80\% training, 10\% validation, and 10\% test data. For Random Forest Regression, the NMSE was 0.27, and Behavioral CNN significantly lowered this to 0.02. To show how this accuracy difference impacts on the performance of the robot, we used both prediction methods in the D* lite path planning algorithm.






\subsection{Evaluation of path planning}

\begin{figure}[htpb]
    \centering
    \includegraphics[width=0.45\textwidth]{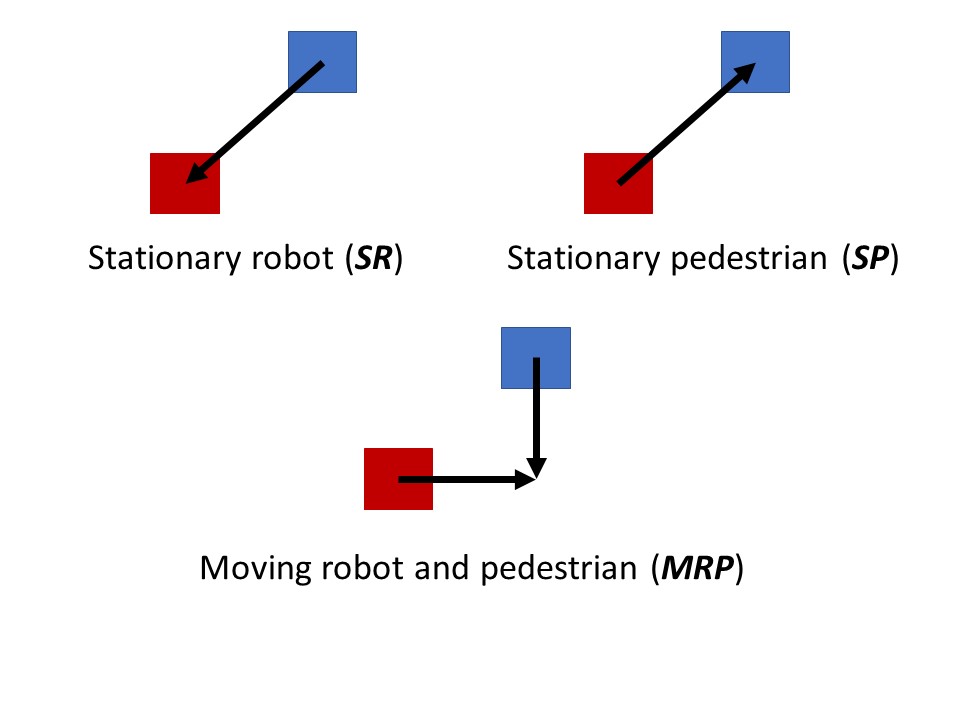}
    \caption{The three types of collisions between the robot (red square) and a pedestrian (blue square).}
    \label{fig:collision_types}
\end{figure}

All path planning algorithms were evaluated in terms of pedestrian safety and time taken to move from the start to the goal position. These metrics were measured as follows:
\begin{itemize}
    \item Number of collisions (i.e. states in which the robot ends up in the same cell as a pedestrian) which were split into 3 groups as shown in Figure~\ref{fig:collision_types}. These represent a robot colliding a stationary pedestrian, a pedestrian colliding with a stationary robot, and a moving robot and a moving pedestrian colliding with each other. 
    \item Relative delay compared to the time of the optimal route with no pedestrians present, as computed by $D = (\frac{\hat{t}}{t} - 1) \times 100\%$, where $t$ is the time taken to move along the optimal route from start to goal, and $\hat{t}$ is the actual time taken.
\end{itemize}

In order to evaluate the path planning approaches, we generated 1000 episodes with random start and goal positions for the robot, and random starting times of the recording data.

\begin{figure}[htpb]
    \centering
    \includegraphics[width=0.45\textwidth]{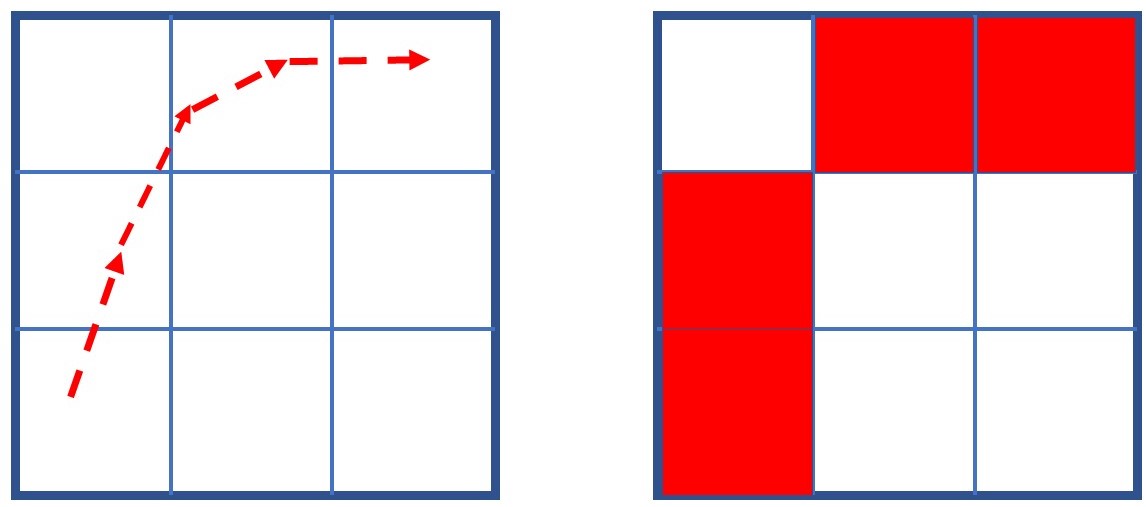}
    \caption{Mapping of vector space movement of SA-CADRL agent to the grid-space representation.}
    \label{fig:route}
\end{figure}

While the SA-CADRL agent moves in a continuous coordinate space, this movement needed to be transformed into the grid-space model of the environment. Figure~\ref{fig:route} shows an example of this mapping. In the first time step the SA-CADRL agent moves one step to the north, and then moves diagonally to the north east. In the third step, however, the agent moves, but does not leave the current grid cell. This means that the agent was stationary for one time step in the grid-space representation. Note that the preferred speed of the SA-CADRL agent was set to match the grid cell size.

Note that the pedestrian movement was recorded without a robot being present, which limits the realism of the simulation to some extent since it ignores the effect that robot movements could influence pedestrian decisions. However, the definition of the three different collision types has been specifically designed to take this limitation into account. Specifically, a moving robot colliding with a moving pedestrian is always an indicator for sub-optimal robot behaviour, no matter whether the pedestrian movement was influenced by the robot or not.

\section{Results}

The results of our evaluation are shown in table~\ref{tab:results}. While the baseline prediction approach may appear to be the most cautious one by assuming that a pedestrian is going to occupy all neighboring squares in the next time step, it caused the most collisions. There were due to some pedestrians moving faster than one square per time step. Increasing the range of squares that are assumed to be occupied to two does not improve the solution, since this causes the robot to stop moving at all. This confirms that a simple over-cautious solution is not feasible in a crowded space. In terms of movement speed, all other techniques demonstrate comparable performance. 

The D* Lite method does not cause any SR collisions because it never is stationary. While the RL approaches do cause a relatively large number of these collisions, these are not indicative of poor robot behaviour, since pedestrians in a real-world situation are unlikely to collide with a stationary robot, and if they do it would not be the robot's fault. SP collisions do not occur for any of the approaches. 

Most importantly, MRP collisions are relatively frequent for the D* Lite methods, corresponding to the accuracy of the chosen prediction method. The RL methods clearly outperform D* Lite on this measure, and our adaptation of SA-CADRL is more than halving the count of MRP collisions to a relatively small number of 3 in 1000. This clearly shows the superiority of Grid-SA-CADRL in this environment. 

\begin{table}[ht]
\caption{Evaluation of different approaches to robots navigation in crowded environment (taken over 1000 paths)}
\begin{tabular}{|l|p{1cm}|p{1cm}|p{1cm}|p{1cm}|p{1cm}|}
\hline
\multirow{3}{*}{Approach} &
\multirow{3}{*}{delay} &
\multicolumn{3}{c|}{collisions} \\ \cline{3-5} 
& &
SR &
SP &
MRP \\ \hline \hline

\multicolumn{5}{|c|}{D* Lite based approach} \\ \hline \hline
Perfect prediction (0.0) & $1.44\%$ & 0 & 0 & 0 \\ \hline
Baseline & $7.57\%$ & 0 & 0 & 143 \\ \hline \hline

RFR (0.27) & $1.39\%$ & 0 & 0 & 67 \\ \hline
Behavior-CNN (0.02) & $1.42\%$ & 0 & 0 & 25 \\ \hline \hline

\multicolumn{5}{|c|}{Deep Reinforcement Learning approach} \\ \hline \hline

SA-CADRL (checkpoint) & $1.57\%$ & 183 & 0 & 8 \\ \hline
Grid-SA-CADRL & $1.46\%$ & 51 & 0 & 3 \\ \hline

\end{tabular}

\label{tab:results}
\end{table}

\section{Conclusions}

In this paper, we have presented a comparative evaluation of various path planning methods for robot navigation through crowds. The evaluation was carried out in a simulation based on real-world recorded pedestrian data in a crowded space. The results show the clear superiority of reinforcement learning in terms of number of collisions with moving pedestrians, while also never colliding with stationary pedestrians. 

\bibliography{bibliography}

\begin{thebibliography}{10}
\providecommand{\url}[1]{#1}
\csname url@samestyle\endcsname
\providecommand{\newblock}{\relax}
\providecommand{\bibinfo}[2]{#2}
\providecommand{\BIBentrySTDinterwordspacing}{\spaceskip=0pt\relax}
\providecommand{\BIBentryALTinterwordstretchfactor}{4}
\providecommand{\BIBentryALTinterwordspacing}{\spaceskip=\fontdimen2\font plus
\BIBentryALTinterwordstretchfactor\fontdimen3\font minus
  \fontdimen4\font\relax}
\providecommand{\BIBforeignlanguage}[2]{{%
\expandafter\ifx\csname l@#1\endcsname\relax
\typeout{** WARNING: IEEEtran.bst: No hyphenation pattern has been}%
\typeout{** loaded for the language `#1'. Using the pattern for}%
\typeout{** the default language instead.}%
\else
\language=\csname l@#1\endcsname
\fi
#2}}
\providecommand{\BIBdecl}{\relax}
\BIBdecl

\bibitem{bai2015intention}
H.~Bai, S.~Cai, N.~Ye, D.~Hsu, and W.~S. Lee, ``Intention-aware online pomdp
  planning for autonomous driving in a crowd,'' in \emph{Robotics and
  Automation (ICRA), 2015 IEEE International Conference on}.\hskip 1em plus
  0.5em minus 0.4em\relax IEEE, 2015, pp. 454--460.

\bibitem{chen2017decentralized}
Y.~F. Chen, M.~Liu, M.~Everett, and J.~P. How, ``Decentralized
  non-communicating multiagent collision avoidance with deep reinforcement
  learning,'' in \emph{Robotics and Automation (ICRA), 2017 IEEE International
  Conference on}.\hskip 1em plus 0.5em minus 0.4em\relax IEEE, 2017, pp.
  285--292.

\bibitem{d-star-lite}
S.~Koenig and M.~Likhachev, ``D\^{}* lite,'' \emph{Aaai/iaai}, vol.~15, 2002.

\bibitem{chen2017socially}
Y.~F. Chen, M.~Everett, M.~Liu, and J.~P. How, ``Socially aware motion planning
  with deep reinforcement learning,'' in \emph{Intelligent Robots and Systems
  (IROS), 2017 IEEE/RSJ International Conference on}.\hskip 1em plus 0.5em
  minus 0.4em\relax IEEE, 2017, pp. 1343--1350.

\bibitem{yi2015understanding}
S.~Yi, H.~Li, and X.~Wang, ``Understanding pedestrian behaviors from stationary
  crowd groups,'' in \emph{Proceedings of the IEEE Conference on Computer
  Vision and Pattern Recognition}, 2015, pp. 3488--3496.

\bibitem{ferrer2014proactive}
G.~Ferrer and A.~Sanfeliu, ``Proactive kinodynamic planning using the extended
  social force model and human motion prediction in urban environments,'' in
  \emph{Intelligent robots and systems (IROS 2014), 2014 IEEE/RSJ international
  conference on}.\hskip 1em plus 0.5em minus 0.4em\relax IEEE, 2014, pp.
  1730--1735.

\bibitem{ferrer2014behavior}
G.~Ferrer~M{\'\i}nguez and A.~Sanfeliu~Cort{\'e}s, ``Behavior estimation for a
  complete framework for human motion prediction in crowded environments,'' in
  \emph{Proceedings of the 2014 IEEE International Conference on Robotics and
  Automation}, 2014, pp. 5940--5945.

\bibitem{yi2016pedestrian}
S.~Yi, H.~Li, and X.~Wang, ``Pedestrian behavior understanding and prediction
  with deep neural networks,'' in \emph{European Conference on Computer
  Vision}.\hskip 1em plus 0.5em minus 0.4em\relax Springer, 2016, pp. 263--279.

\bibitem{sisbot2007human}
E.~A. Sisbot, L.~F. Marin-Urias, R.~Alami, and T.~Simeon, ``A human aware
  mobile robot motion planner,'' \emph{IEEE Transactions on Robotics}, vol.~23,
  no.~5, pp. 874--883, 2007.

\bibitem{reynolds1999steering}
C.~W. Reynolds, ``Steering behaviors for autonomous characters,'' in \emph{Game
  developers conference}, vol. 1999.\hskip 1em plus 0.5em minus 0.4em\relax
  Citeseer, 1999, pp. 763--782.

\bibitem{svenstrup2010trajectory}
M.~Svenstrup, T.~Bak, and H.~J. Andersen, ``Trajectory planning for robots in
  dynamic human environments,'' in \emph{Intelligent robots and systems (IROS),
  2010 IEEE/RSJ international conference on}.\hskip 1em plus 0.5em minus
  0.4em\relax IEEE, 2010, pp. 4293--4298.

\bibitem{kretzschmar2016socially}
H.~Kretzschmar, M.~Spies, C.~Sprunk, and W.~Burgard, ``Socially compliant
  mobile robot navigation via inverse reinforcement learning,'' \emph{The
  International Journal of Robotics Research}, vol.~35, no.~11, pp. 1289--1307,
  2016.

\bibitem{liaw2002classification}
A.~Liaw, M.~Wiener \emph{et~al.}, ``Classification and regression by
  randomforest,'' \emph{R news}, vol.~2, no.~3, pp. 18--22, 2002.

\end{thebibliography}
\bibliographystyle{IEEEtran}

\end{document}